\documentclass[runningheads]{llncs}
\usepackage[T1]{fontenc}
\usepackage{graphicx}
\usepackage{amsmath, amssymb}
\usepackage{booktabs}
\usepackage{multirow}
\usepackage{cite}
\usepackage{url}
\usepackage{float}
\usepackage[utf8]{inputenc}
\usepackage{xcolor}
\usepackage{pifont} 
\usepackage{marvosym}

\begin{document}
\title{MATHENA: Mamba-based Architectural Tooth Hierarchical Estimator and Holistic Evaluation Network for Anatomy}
\titlerunning{MATHENA: Mamba-based Tooth Estimator and Evaluation Network}
%

\author{Kyeonghun Kim\inst{1} \and
Jaehyung Park\inst{2} \and
Youngung Han\inst{1, 2} \and
Anna Jung\inst{2} \and\\
Seongbin Park\inst{2} \and
Sumin Lee\inst{2} \and
Jiwon Yang\inst{2} \and
Jiyoon Han\inst{2} \and
Subeen Lee\inst{2} \and\\
Junsu Lim\inst{1, 3} \and
Hyunsu Go\inst{2} \and
Eunseob Choi\inst{4} \and
Hyeonseok Jung\inst{2} \and
Soo Yong Kim\inst{2} \and\\
Woo Kyoung Jeong\inst{5} \and
Won Jae Lee\inst{6} \and
Pa Hong\inst{6} \and
Hyuk-Jae Lee\inst{2}\\
Ken Ying-Kai Liao\inst{7} \and
Nam-Joon Kim\inst{2}\textsuperscript{(\Letter)}
}
\authorrunning{K. Kim et al.}
\institute{OUTTA, Seoul, South Korea\\
\email{kyeonghun.kim@outta.ai} \and
Seoul National University, Seoul, South Korea\\
\email{knj01@snu.ac.kr} \and
Sangmyung University, Seoul, South Korea \and
Gwangju Institute of Science and Technology, Gwangju, South Korea \and
Samsung Medical Center, Sungkyunkwan University, Seoul, South Korea \and
Samsung Changwon Hospital, Sungkyunkwan University, Changwon, South Korea \and
NVIDIA AI Technology Center, Taipei, Taiwan}

\maketitle

\begin{abstract}
\sloppy

Dental diagnosis from Orthopantomograms (OPGs) requires coordination of tooth detection, caries segmentation (CarSeg), anomaly detection (AD), and dental developmental staging (DDS). We propose Mamba-based Architectural Tooth Hierarchical Estimator and Holistic Evaluation Network for Anatomy (\textbf{MATHENA}), a unified framework leveraging Mamba's linear-complexity State Space Models (SSM) to address all four tasks. \textbf{MATHENA} integrates \textbf{MATHE}, a multi-resolution SSM-driven detector with four-directional Vision State Space (VSS) blocks for $O(N)$ global context modeling, generating per-tooth crops. These crops are processed by \textbf{HENA}, a lightweight Mamba-UNet with a triple-head architecture and Global Context State Token (GCST). In the triple-head architecture, CarSeg is first trained as an upstream task to establish shared representations, which are then frozen and reused for downstream AD fine-tuning and DDS classification via linear probing, enabling stable, efficient learning. We also curate \textbf{PARTHENON}, a benchmark comprising 15,062 annotated instances from ten datasets. \textbf{MATHENA} achieves 93.78\% $\mathrm{mAP}_{50}$ in tooth detection, 90.11\% Dice for CarSeg, 88.35\% for AD, and 72.40\% ACC for DDS.

%
%

\end{abstract}

\section{Introduction}
Orthopantomogram (OPG) is the widely used dental radiographs that provide a comprehensive overview of dental arches, maxillary and mandibular bones, and surrounding anatomical structures~\cite{izzetti2021basic}. Clinical diagnosis requires integrated evaluation of tooth detection, caries segmentation (CarSeg), anomaly detection (AD), and dental developmental staging (DDS)~\cite{white2014oral,demirjian1973new}. While clinical dentistry distinguishes congenital anomalies from acquired pathoses~\cite{white2014oral}, computer vision defines anomaly detection more broadly as identifying morphological deviation from normal distributions~\cite{pang2021deep}. Adopting this computational perspective, we formulate AD as a generalized pixel-wise segmentation task. By unifying dental lesions into a single anomalous class, our approach integrates AD with other diagnostic tasks and emphasizes pathological localization.

Current deep learning models often address these tasks in isolation: CNNs lack global context~\cite{faster_rcnn}, while Transformers incur quadratic computational overhead~\cite{liu2021swin,rtdetr}. To overcome these limitations, we propose MATHENA, a unified framework reflecting the clinical coarse-to-fine workflow. MATHENA incorporates Mamba~\cite{gu2024mamba, Xin_SegMamba_MICCAI2024, Vision_mamba}--a selective State Space Model with linear $O(N)$ complexity and global receptive fields--into MATHE for tooth detection and HENA for per-tooth multi-task analysis.

\begin{figure}[t]
\centering
\includegraphics[width=\textwidth]{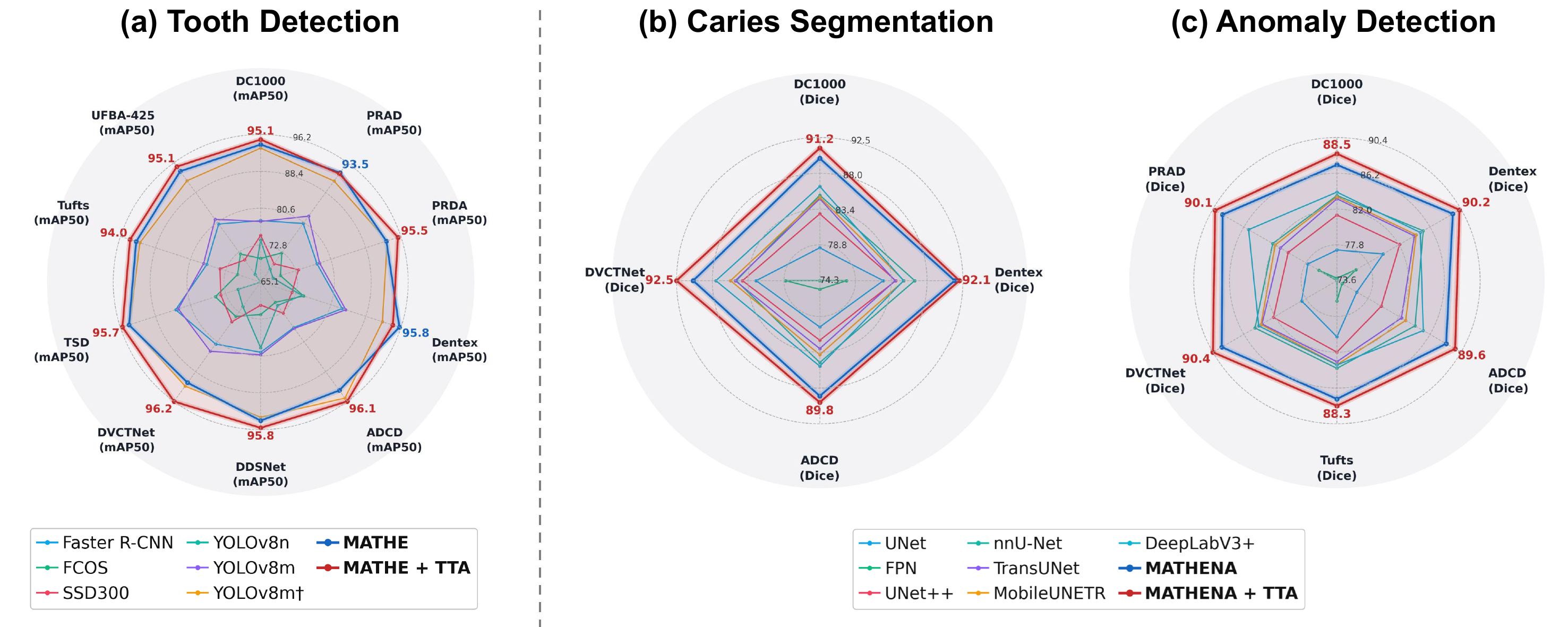}
\caption{Quantitative results on PARTHENON: (a) MATHE variants outperform baseline models in tooth detection (mAP50); (b, c) MATHENA variants show superior Dice scores in CarSeg and AD. ($\dagger$: baseline enhanced with P2, BiFPN, and WIoU).}

\label{fig1}
\end{figure}

To address fragmentation across existing datasets, we curate PARTHENON, a large-scale benchmark unifying ten datasets into 15,062 annotated instances. As shown in Fig.~\ref{fig1}, MATHENA consistently outperforms existing baselines in tooth detection, CarSeg, and AD across the individual datasets comprising PARTHENON. Our main contributions can be summarized as follows:
\begin{itemize}
     \item We propose MATHENA, a unified framework for tooth detection, CarSeg, AD, and DDS.
      \item We integrate directional Vision State Space (VSS) blocks to achieve $O(N)$ global context modeling without Transformer overhead.
       \item We enable per-tooth multi-task prediction via a novel Global Context State Token (GCST) mechanism and triple-head design.
\end{itemize}

\section{Dataset}

\subsection{PARTHENON Dataset}

As shown in Table~\ref{tab:parthenon}, PARTHENON aggregates ten dental datasets (8 panoramic, 2 periapical; 15,062 instances) with annotations spanning 14 original diagnostic categories, which are merged into task-specific binary labels (Sec.~\ref{sec:dataprep}). Annotations consist of tooth-level bounding boxes, either manually annotated or generated from existing segmentation masks. For subsets with developmental metadata, dental maturity is categorized using the Demirjian method (A-H)~\cite{demirjian1973new}. Each dataset supports one or more tasks summarized in Table~\ref{tab:parthenon}.

\newcommand{\taskA}{\ding{172}} 
\newcommand{\taskB}{\ding{173}} 
\newcommand{\taskC}{\ding{174}} 
\newcommand{\taskD}{\ding{175}} 

\begin{table}[h]
\centering
\caption{PARTHENON composition: \ding{172}-\ding{175} denote tooth detection, Caries Segmentation (CarSeg), Anomaly Detection (AD), and Dental Developmental Staging (DDS).}

\label{tab:parthenon}
\small
\begin{tabular}{lcccccc}
\toprule
\textbf{No.} & \textbf{Dataset} & \textbf{Image} & \textbf{Annotation} & \textbf{CarSeg Mask} & \textbf{AD Mask} & \textbf{Task} \\
\midrule
D1 & DC1000  \cite{dc1000}         & 597    & 597   & 591   & 591 & \textbf{\taskB\ \taskC}\\
D2 & PRAD \cite{zhou2025prad}            & 5,000  & 5,000    & ---   & 669 & \textbf{\taskA\ \taskC}\\
D3 & PRDA \cite{fatima2023deep}  & 532 & 532  & ---  & 516 & \textbf{\taskC} \\
D4 & Dentex \cite{hamamci2023dentex}   & 2,326   & 2,326   & 694   & 724 & \textbf{\taskB\ \taskC}\\
D5 & ADCD \cite{adcd}       & 1,808    & 1,808   & 1,068  & 1,559 & \textbf{\taskA\ \taskB\ \taskC} \\
D6 & DDSNet  \cite{wang2025ddsnet}      & 380    & 380   & ---   & --- & \textbf{\taskA\ \taskD} \\
D7 & DVCTNet \cite{dvctnet}          & 2,000    & 2,000   & 1,771   & 1,771 & \textbf{\taskB\ \taskC} \\
D8 & TSD \cite{tsd}      & 994    & 994   & 611   & 611 & \textbf{\taskA\ \taskB\ \taskC} \\
D9 & Tufts \cite{tufts2019}      & 1,000 & 1,000   & ---   & --- & \textbf{\taskA} \\
D10 & UFBA-425 \cite{ufba425}         & 425    & 425   & ---   & --- & \textbf{\taskA} \\
\midrule
\textbf{Total}  &            & \textbf{15,062} & \textbf{15,062} & \textbf{4,735} & \textbf{6,411} & \\
\bottomrule
\end{tabular}
\end{table}

\subsection{Data Preprocessing}
\label{sec:dataprep}

\subsubsection{Semi-Supervised Pseudo-Label Generation.} Ground-truth bounding boxes for tooth detection are available for PARTHENON subsets D2, D5, D6, D8, D9, D10. RT-DETR-L, trained on these subsets, achieves 93.7 $\mathrm{mAP}_{50}$ and serves as the teacher. We apply it to the other datasets in a semi-supervised framework~\cite{liu2021unbiased} to generate pseudo-ground-truth bounding boxes~\cite{yang2022survey,rtdetr}.

\subsubsection{Pseudo-Label Quality Filtering.}
Teacher-generated bounding boxes are filtered using a confidence threshold and NMS. Anatomically implausible predictions are rejected by Mahalanobis distance-based outlier detection~\cite{mueller2025mahalanobis++}. Each box is mapped to normalized spatial features $v{=} [c_x/W,\;c_y/H,\; \log(w/W),\; \log(h/H)]^\top$, and predictions whose squared distance exceeds $\chi^2_4$ threshold at $p < 0.001$ are discarded~\cite{tabachnick2007using, rousseeuw1990unmasking}. The resulting bounding boxes are used to train MATHE.

\subsubsection{Label Merging.}
For CarSeg, the multi-stage annotations in D1, D4, D5, D7, and D8 are collapsed into a binary mask (caries vs.\ background) by mapping all non-zero pixel classes to~1. For AD, per-tooth labels in D1, D2, D3, D4, D5, and D7 are unified into a binary normal/anomalous label.

\subsubsection{Cropping and Augmentation.}
Cropped image-mask pairs are extracted from the OPG and corresponding mask for HENA training. Offline augmentations--random rotations and horizontal flip--are applied to each instance.

\section{Methodology: The MATHENA Framework}

\begin{figure}[t]
\centering
\includegraphics[width=\textwidth]{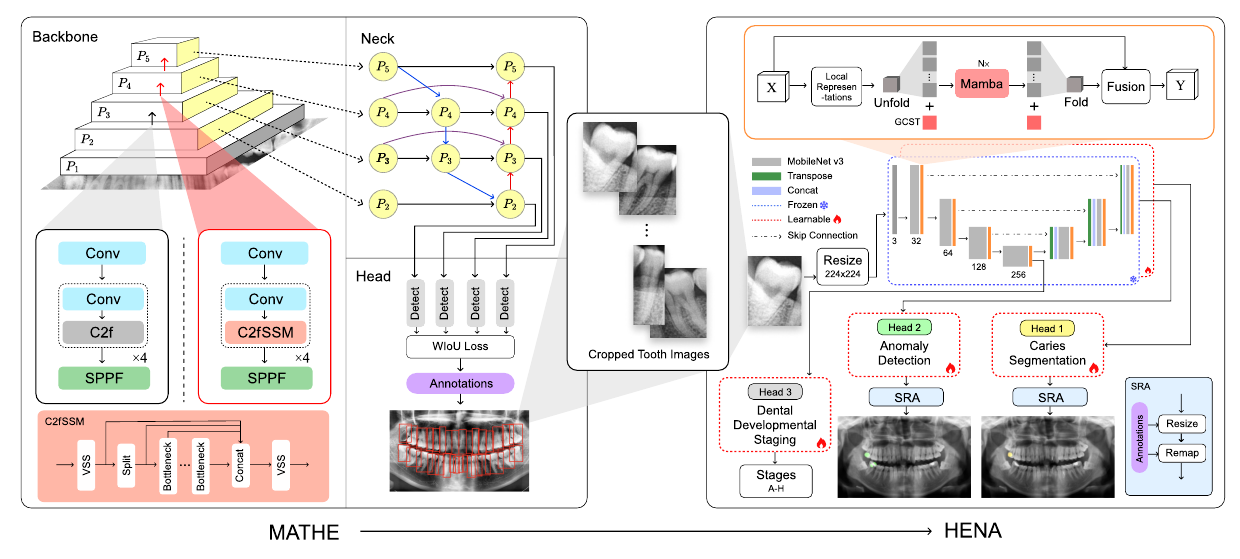}
\caption{MATHENA architecture: \textbf{Left} shows MATHE backbone, BiFPN, and detection head; \textbf{Right} depicts HENA encoder-decoder with GCST skip fusion. SRA (Spatial Re-Alignment) remaps the output mask to the original OPG for visualization.}

\label{fig2}
\end{figure}

We propose MATHENA, a unified framework motivated by the coarse-to-fine clinical review process, as illustrated in Fig.~\ref{fig2}. Our framework is twofold: MATHE for tooth detection (Sec.~\ref{sec:mathe}) and HENA for multi-task analysis, including CarSeg, AD, and DDS (Sec.~\ref{sec:hena}).

\subsection{MATHE: Mamba-based Architectural Tooth Hierarchical Estimator}
\label{sec:mathe}

Given an OPG image, MATHE extracts multi-scale features, fuses them across resolutions, and outputs per-tooth bounding boxes.

\subsubsection{Hybrid CNN-SSM Backbone.}
Early stages ($P_2$, $P_3$) use standard convolutions to capture local features. Deeper stages ($P_4$, $P_5$) replace bottlenecks with C2fSSM blocks containing VSS units that perform four-directional selective scanning. This allocates $O(N)$ global context modeling at semantically rich stages while retaining efficient convolutions at high resolution.

\subsubsection{Bidirectional Feature Pyramid and Head.}
Multi-scale features from four backbone stages are fused through a Bidirectional Feature Pyramid Network (BiFPN)~\cite{tan2020efficientdet} with learnable per-node fusion weights. Unlike standard FPN, BiFPN adds a bottom-up pathway and weighted feature combination at each fusion node:
\[
P_i^{out}{=}\frac{\sum_j w_j \cdot P_j^{in}}{\sum_j w_j + \epsilon},
\]
where $w_j$ are learnable scalar weights and $\epsilon$ ensures numerical stability. We include $P_2$ (stride 4) to improve detection of small periapical structures lost at coarser scales~\cite{yu2023enhanced}. The head uses decoupled convolutional towers for box regression at each pyramid level and is optimized with Wise-IoU (WIoU)~\cite{tong2023wise}, which dynamically adjusts gradients based on box quality.

\subsection{HENA: Holistic Evaluation Network for Anatomy}
\label{sec:hena}

Each detected tooth region with its paired mask is cropped, resized to $224\times224$, and passed through HENA's pipeline before being routed to task-specific heads.

\subsubsection{Lightweight Encoder.}
We employ a lightweight U-shaped encoder-decoder inspired by MobileUNETR~\cite{perera2024mobileunetr}, where we replace all Transformer blocks with Mamba VSS blocks ~\cite{ Xin_SegMamba_MICCAI2024} to achieve $O(N)$ complexity for intra-tooth dependency modeling. The encoder consists of depthwise-separable convolution (DWSep) blocks~\cite{howard2017mobilenets} to progressively downsample spatial resolution while expanding channel depth.

\subsubsection{Mamba Bottleneck with GCST.}
At the $28{\times}28$ bottleneck, the feature map $F_{bot} \in \mathbb{R}^{B{\times}256{\times}28{\times} 28}$ is flattened to a sequence $X \in \mathbb{R}^{B{\times}784{\times}256}$. We prepend a learnable global context token $T_g \in \mathbb{R}^{1{\times}256}$ (initialized to zeros) to form $X'{=}[T_g; X] \in \mathbb{R}^{B{\times}785{\times}256}$. A VSS block processes $X'$ with four-directional selective scanning, producing $H{=}\mathrm{VSS}(X') \in \mathbb{R}^{B{\times} 785{\times}256}$. We extract the global token state $h_g{=}H_0 \in \mathbb{R}^{C}$ and the spatial hidden states $Z{=}H_{1:L} \in \mathbb{R}^{L{\times}C}$, then broadcast-add the global context to all spatial positions:
$$
Y_{out}{=}Z + \mathbf{1}_L h_g^\top,
$$
where $h_g$ acts as a global context aggregator. It accumulates holistic tooth-level semantics through Mamba's linear recurrence, providing dense spatial modulation at an efficient $O(N)$ cost. The modulated features are then reshaped to $\mathbb{R}^{B{\times}256 {\times}28{\times}28}$ for the decoder.

\subsubsection{Decoder with GCST Skip Fusion.}
The decoder progressively restores spatial resolution through transposed convolutions with skip connections. At each decoder level $s \in \{1, 2, 3\}$, the skip features $S_s \in \mathbb{R}^{B{\times}C_s{\times}H_s{\times}W_s}$ are modulated by GCST before fusion. The mechanism flattens $S_s$ to $X_s \in \mathbb{R}^{B{\times}L_s {\times}C_s}$ ($L_s{=}H_s{ \times}W_s$), prepends a learnable scale token $T_s$, and applies a lightweight Mamba block to $[T_s; X_s]$. The output token state $h_s$ is projected to Feature-wise Linear Modulation (FiLM)~\cite{perez2018film} parameters:
\[
(\gamma_s, \beta_s){=}\psi(h_s), \qquad
\hat{S}_s{=}\gamma_s \odot S_s + \beta_s,
\]
where $\gamma_s, \beta_s \in \mathbb{R}^{C_s}$ are broadcast
spatially. This replaces the Transformer-based cross-attention in
MobileUNETR with an $O(L_s)$ alternative that captures cross-scale
dependencies through Mamba's selective recurrence.

\begin{table}[t]
\centering
\caption{Tooth detection on PARTHENON: $^\star$baseline YOLOv8; $^\dagger$enhanced with P2, BiFPN, WIoU.}
\label{tab:_results}
\small
\begin{tabular}{lcccc}
\toprule
\textbf{Method} & \textbf{Backbone} & \textbf{mAP$_{50}$(\%)} & \textbf{mAP$_{75}$(\%)} & \textbf{mAP$_{50:95}$(\%)} \\
\midrule
RetinaNet            & ResNet-50        & 71.01 & 61.81 & 51.83 \\
YOLOv8n$^\star$      & CSPDarkNet(C2f) & 71.30 & 66.30 & 61.90 \\
FCOS                 & ResNet-50        & 71.86 & 62.20 & 53.20 \\
SSD300               & VGG-16           & 72.77 & 65.60 & 58.80 \\
YOLOv8s$^\star$      & CSPDarkNet(C2f) & 73.66 & 70.05 & 65.02 \\
Faster R-CNN         & MobileNetV3      & 80.08 & 66.56 & 49.30 \\
Faster R-CNN         & ResNet-50        & 80.32 & 68.97 & 50.72 \\
YOLOv8m$^\star$      & CSPDarkNet(C2f) & 80.32 & 75.71 & 67.80 \\
YOLOv8n$^\dagger$    & CSPDarkNet(C2f) & 90.11 & 85.30 & 77.90 \\
YOLOv8s$^\dagger$    & CSPDarkNet(C2f) & 92.24 & 89.89 & 79.32 \\
YOLOv8m$^\dagger$    & CSPDarkNet(C2f) & 92.78 & 88.77 & 79.91 \\
\midrule
\textbf{MATHE}       & Mamba-SSM        & 93.78 & 91.89 & 81.32 \\
\textbf{MATHE + TTA} & Mamba-SSM        & \textbf{94.89} & \textbf{92.94} & \textbf{83.45} \\
\bottomrule
\end{tabular}
\end{table}

\subsubsection{Triple-Head Multi-Task Learning and Training Strategy.}
HENA employs a shared encoder-decoder backbone with three task-specific heads, optimized via sequential transfer learning. First, the entire network is trained on the upstream CarSeg task to establish robust dental representations. Next, treating AD and DDS as downstream tasks, we freeze the shared backbone to efficiently transfer these learned features. The AD head is attached to the decoder, while the DDS head is applied to the encoder's bottleneck ($GAP(F_{bot})$) and fine-tuned via linear probing. At inference, the predicted stage is $\hat{s}{=}\textstyle\sum_{j{=}1}^{8}\mathbf{1}[\sigma(\hat{y}_j){>}0.5]$, mapped to A-H.

Freezing the common backbone for downstream tasks minimizes computational overhead. Compared to a fully learnable setup (90.03\% CarSeg Dice), our frozen sequential approach maintained 90.11\% Dice while reducing training and inference times by 3.5${\times}$ and 1.4${\times}$, respectively.

\subsection{Loss Functions}

The MATHE detector and HENA analyzer are trained with:
\begin{equation}
    \mathcal{L}_{MATHE} = \lambda_{wiou}\mathcal{L}_{WIoU}(B, \hat{B}) + \lambda_{l1}\mathcal{L}_{L1}(B, \hat{B}) + \lambda_{dfl}\mathcal{L}_{DFL}(B, \hat{B})
\end{equation}
\begin{equation}
    \mathcal{L}_{HENA}{=}\mathcal{L}_{Dice}(S, \hat{S}) + \mathcal{L}_{Dice}(A, \hat{A}) + \mathcal{L}_{Ord}(Y_{stg}, \hat{Y}_{stg})
\end{equation}
where $\mathcal{L}_{WIoU}$~\cite{tong2023wise} is the Wise-IoU loss, $\mathcal{L}_{L1}$ is the bounding box regression loss, $\mathcal{L}_{DFL}$ is the Distribution Focal Loss addressing the bounding box coordinate distribution, $\mathcal{L}_{Dice}$~\cite{dice_loss} supervises CarSeg and AD, and $\mathcal{L}_{Ord}$~\cite{ordinal_reg} is the cumulative ordinal loss with $K_{stg}=8$ stages and $K_{stg}-1$ binary thresholds.

\section{Experiments}

\subsection{Implementation Details}

MATHENA was implemented in PyTorch on an NVIDIA A100 GPU. MATHE builds on YOLOv8m~\cite{yaseen2025yolov8}, replacing C2f blocks at $P_4$ and $P_5$ with C2fSSM, FPN with BiFPN, and IoU with WIoU. Training was performed for 100 epochs at $1024{\times}512$ using AdamW ($lr{=}1{\times}10^{-4}$, weight decay $5{\times}10^{-2}$) with linear warmup and cosine annealing. HENA was trained for 150 epochs on $224{\times}224$ crops with batch size 16. TTA applies horizontal flip with NMS (threshold 0.5), improving MATHE from 93.78\% to 94.89\% mAP$_{50}$.

\subsection{Comparative Analysis}

\begin{figure}[!t]
\centering
\includegraphics[width=\textwidth]{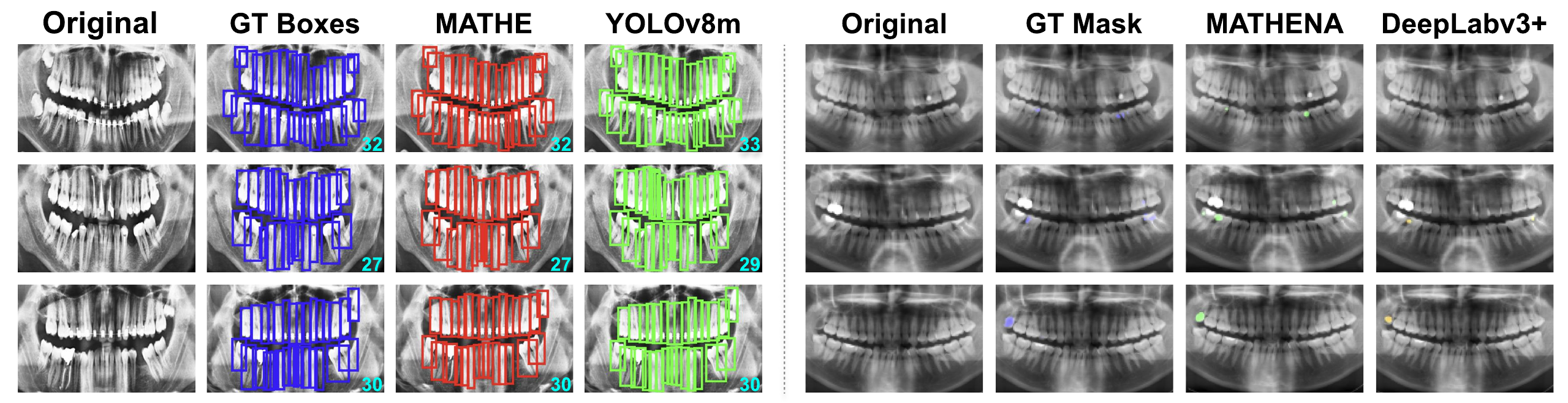}
\caption{Visual comparison:  \textbf{Left} shows Tooth Detection; \textbf{Right} depicts CarSeg.}\label{fig3}
\end{figure}

We compared MATHE and MATHENA with standard object detectors and leading segmentation models on the PARTHENON test set. Quantitative performance is summarized in Table~\ref{tab:_results} and ~\ref{tab:parthenon_multitask}, with a visual comparison in Fig.~\ref{fig3}.

\begin{table}[t]
\centering

\caption{Multi-task performance on PARTHENON: CarSeg and AD averaged across subsets; DDS on DDSNet.}

\label{tab:parthenon_multitask}
\setlength{\tabcolsep}{3pt}
\begin{tabular}{lcccccc}
\toprule
 & \multicolumn{2}{c}{\textbf{CarSeg}} & \multicolumn{2}{c}{\textbf{AD}} & \multicolumn{2}{c}{\textbf{DDS}} \\
\cmidrule(lr){2-3} \cmidrule(lr){4-5} \cmidrule(lr){6-7}
\textbf{Method} & Dice (\%) & IoU (\%) & Dice (\%) & IoU (\%) & ACC (\%) & F1 (\%) \\
\midrule
FPN            & 76.41 & 63.93 & 74.74 & 61.29 & -- & -- \\
MAnet          & 77.51 & 64.48 & 75.29 & 62.03 & -- & -- \\
PSPNet         & 79.21 & 66.12 & 76.55 & 63.37 & -- & -- \\
UNet           & 80.81 & 67.24 & 78.16 & 64.65 & -- & -- \\
Linknet        & 82.21 & 68.75 & 79.87 & 66.14 & -- & -- \\
UNet++         & 83.11 & 69.57 & 81.30 & 67.83 & -- & -- \\
SEResUNet      & 83.28 & 69.89 & 81.65 & 68.06 & -- & -- \\
UNet3+         & 84.12 & 70.64 & 82.38 & 68.92 & -- & -- \\
TransUNet      & 84.64 & 71.13 & 83.03 & 69.59 & -- & -- \\
MobileUNETR    & 84.82 & 71.47 & 83.40 & 69.90 & -- & -- \\
nnU-Net        & 85.23 & 72.12 & 83.91 & 69.87 & -- & -- \\
DeepLabv3+     & 85.91 & 72.83 & 84.48 & 70.74 & -- & -- \\
\midrule
\textbf{MATHENA}        & \textbf{90.11} & \textbf{76.94} & \textbf{88.35} & \textbf{74.77} & \textbf{72.40} & \textbf{70.10} \\
\textbf{MATHENA + TTA}  & \textbf{91.31} & \textbf{78.05} & \textbf{89.59} & \textbf{75.92} & \textbf{74.10} & \textbf{72.30} \\
\bottomrule
\end{tabular}
\end{table}

\textbf{Tooth Detection.}
MATHE achieves 93.78\% mAP$_{50}$, improving over baseline YOLOv8 configurations. TTA, merging multi-view predictions via NMS, increases mAP$_{50}$ to 94.89\%.

\textbf{Multi-task Performance.} MATHENA reaches 90.11\% Dice for CarSeg and 88.35\% for AD, outperforming baselines including DeepLabv3+ (85.91\%, 84.48\%). As shown in Fig.~\ref{fig3}, MATHENA provides precise tooth detection and segmentation across tasks.

\textbf{DDS.} The DDS head achieves 72.40\% ACC and 70.10\% F1 on DDSNet.

\subsection{Ablation Study}

Table~\ref{tab:ablation} validates the architectural design of MATHENA across all four tasks. Removing GCST from MATHE drops detection performance by 4.36\% mAP$_{50}$, indicating that global context is critical for resolving crowded dentition. Replacing BiFPN with a standard FPN reduces detection by 3.14\% mAP$_{50}$, confirming the necessity of bidirectional multi-scale fusion for accurate tooth detection. For per-tooth analysis, removing the GCST skip fusion degrades CarSeg by 3.15\% Dice, highlighting its essential role in cross-scale spatial modulation. Replacing the Mamba bottleneck with standard convolutions degrades CarSeg by 2.36\% Dice and DDS by 2.45\% ACC, proving the efficacy of linear recurrence for holistic feature extraction. Finally, replacing Mamba blocks with Vision Transformers universally underperforms, demonstrating Mamba's superior architectural stability and efficiency in capturing intra-tooth dependencies without quadratic computational overhead.



\begin{table}[!t]
\centering
\caption{Ablation study on PARTHENON: impact of each component across all tasks.}\label{tab:ablation}
\small
\begin{tabular}{lcccc}
\toprule
\textbf{Configuration} & \textbf{Detection (\%)} & \textbf{CarSeg (\%)} & \textbf{AD (\%)} & \textbf{DDS (\%)} \\
\midrule
\textbf{Full MATHENA}                      & \textbf{93.78} & \textbf{90.11} & \textbf{88.35} & \textbf{72.40} \\
\quad w/o GCST in MATHE               & 89.42  & 90.05 & 88.29 & 72.36 \\
\quad w/o Mamba in HENA BN    & 93.74 & 87.75  & 85.99 & 69.95 \\
\quad w/o GCST skip fusion            & 93.75 & 86.96  & 85.20 & 72.32 \\
\quad w/ standard FPN(no BiFPN)      & 90.64 & 90.03 & 88.27 & 72.25 \\
\quad w/ Transformer(vs. Mamba)      & 92.13 & 88.66 & 86.90 & 70.94 \\
\bottomrule
\end{tabular}
\end{table}

\section{Conclusion}
We present MATHENA, a unified framework integrating MATHE and HENA for tooth detection and multi-task analysis. We also introduce PARTHENON, a benchmark unifying ten datasets under a common schema. The clinically motivated coarse-to-fine pipeline addresses tooth detection, CarSeg, AD, and DDS. Experiments demonstrate that MATHENA provides a structured baseline for multi-task panoramic analysis.

%
%
\bibliographystyle{splncs04}
\bibliography{references}

\end{document}